%% file: main.tex
\documentclass[runningheads]{llncs}

\usepackage[T1]{fontenc}

\usepackage{graphicx}

\usepackage{amsmath, amsfonts, amssymb}
\usepackage{xspace}
\usepackage{stmaryrd}
\usepackage{hyperref}
\usepackage[T1]{fontenc}
\usepackage[utf8]{inputenc}
\usepackage{graphicx}
\usepackage{listings}
\usepackage{subcaption}
\usepackage{xcolor}
\usepackage{float}
\usepackage{lstautogobble}
\usepackage{color}
\usepackage{proof}
\usepackage{multicol}
\usepackage{mathpartir}
\usepackage{todonotes}
\usepackage{cleveref}
\usepackage{mathpartir}
\usepackage{todonotes}
\usepackage{wrapfig}
\usepackage{mathtools}
\usepackage{tabularray}
\UseTblrLibrary{booktabs}
\usepackage{orcidlink}
\usepackage{marvosym}
\input{macros.tex}

\usepackage{tikz}
\usetikzlibrary{backgrounds,calc,positioning}
\usepackage{csvsimple}

\definecolor{bluekeywords}{rgb}{0.13, 0.13, 1}
\definecolor{greentypes}{rgb}{0, 0.5, 0}
\definecolor{inferedgreentypes}{rgb}{1.0, 0.2, 0}
\definecolor{orangecomments}{rgb}{1, 0.5, 0.1}
\definecolor{redstrings}{RGB}{171, 114, 2}
\definecolor{graynumbers}{rgb}{0.5, 0.5, 0.5}
\definecolor{goldcomments}{rgb}{0.6, 0.4, 0.08}

\lstdefinelanguage{Lola}{
  keywords=[0]{input, output, trigger, constant, import, spawn, eval, close, with, when},
  moredelim=**[is][\transparent{0.6}]{?}{?},
  moredelim=**[is][\color{greentypes}@]{@}{@},
  keywordstyle=[0]\bfseries\color{bluekeywords},
  keywords=[1]{if, then, else, aggregate, defaults, offset, last, by, or, to, sin, cos, abs, hold, over, using, over_instances},
  keywords=[2]{Variable, String, Int, Int64, UInt, UInt64, Bool, Float32, Float64, Float},
  keywordstyle=[2]\color{greentypes},
  sensitive=false,
  comment=[l]{//},
  morecomment=[s]{/*}{*/},
  morestring=[b]',
  morestring=[b]",
  literate={\\@}{@}1
}
\makeatletter
\newcommand\extrafootertext[1]{%
    \bgroup
    \renewcommand\thefootnote{\fnsymbol{footnote}}%
    \renewcommand\thempfootnote{\fnsymbol{mpfootnote}}%
    \footnotetext[0]{#1}%
    \egroup
}

\begin{document} 

\title{Stream-Based Monitoring of Algorithmic Fairness}
\author{
  Jan Baumeister\textsuperscript{(\Letter)}\orcidlink{0000-0002-8891-7483}
  \and Bernd Finkbeiner\,\orcidlink{0000-0002-4280-8441}
  \and Frederik Scheerer\,\orcidlink{0009-0007-8115-0359}
  \and\\ Julian Siber\,\orcidlink{0000-0003-0842-0029}
  \and Tobias Wagenpfeil\,\orcidlink{0009-0005-2680-1150}
}
\authorrunning{J.\ Baumeister et~al.}

\institute{CISPA Helmholtz Center for Information Security, Saarbrücken, Germany
\email{\{jan.baumeister, finkbeiner, frederik.scheerer, julian.siber, tobias.wagenpfeil\}@cispa.de}
}

\maketitle

\begin{abstract}
Automatic decision and prediction systems are increasingly deployed in applications where they significantly impact the livelihood of people, such as for predicting the creditworthiness of loan applicants or the recidivism risk of defendants. These applications have given rise to a new class of \emph{algorithmic-fairness} specifications that require the systems to decide and predict without bias against social groups. Verifying these specifications statically is often out of reach for realistic systems, since the systems may, e.g., employ complex learning components, and reason over a large input space. In this paper, we therefore propose stream-based monitoring as a solution for verifying the algorithmic fairness of decision and prediction systems at runtime. Concretely, we present a principled way to formalize algorithmic fairness over temporal data streams in the specification language RTLola and demonstrate the efficacy of this approach on a number of benchmarks. Besides synthetic scenarios that particularly highlight its efficiency on streams with a scaling amount of data, we notably evaluate the monitor on real-world data from the recidivism prediction tool COMPAS.

\end{abstract}

\section{Introduction}

Machine learning is used to automate an increasing number of critical decisions pertaining to people's opportunities in areas such as loan or job application~\cite{Amazon}, healthcare~\cite{Davenport}, and criminal sentencing~\cite{ProPublica16}. It is of vital interest that these decision and prediction systems adhere to societies' shared values and, hence, they should in particular not discriminate against members of protected social groups, e.g., based on attributes such as gender or perceived ethnicity. Since the machine-learned systems are trained from historical data, they often inherit the historical human bias present in these data sets. So far, this usually was revealed by posterior analyses after the systems have been deployed for years~\cite{COMPAS,Pessach2023} and hence have already produced harmful results. In this paper, we propose stream-based monitoring of algorithmic fairness properties as a way to alleviate this situation, and to significantly reduce the impact unfair decisions have on people that are subject to learned decision and prediction systems. Unlike static verification of these systems, which is often intractable due to their complex learning components and large input space, monitors are lightweight and can be deployed alongside the machine-learned systems, raising awareness once they are sufficiently sure of unfair behavior. While this does not avert all decisions made by an unfair system, we show empirically that it can still significantly reduce the number of decisions made by an unfair system by alerting practicioners early.

\subsection{Motivating Example}\label{sec:motivation}

\begin{wrapfigure}{R}{0.56\linewidth}
\small
	\vspace{-13.75mm}
	\centering 
         \def\arraystretch{1.3}
		\setlength\tabcolsep{1mm}
		\begin{tabular}{cccccc}
			\toprule
			\texttt{date} & \texttt{event} & \texttt{id} & \texttt{group} & \texttt{risk} & \texttt{\ldots}\\
			\midrule
			\texttt{2013-01-02}  & \texttt{SCREEN} & \texttt{0} & \texttt{A} & \texttt{HIGH} & \texttt{\ldots}\\
			\texttt{2013-01-02} & \texttt{SCREEN} & \texttt{1} & \texttt{B} & \texttt{LOW} & \texttt{\ldots}\\
			\texttt{2013-01-03} & \texttt{RECID.} & \texttt{1} & \texttt{-} & \texttt{-} & \texttt{\ldots}\\
			\texttt{2013-01-03} & \texttt{SCREEN} & \texttt{2} & \texttt{B} & \texttt{HIGH} & \texttt{\ldots}\\
			\texttt{\ldots} & \texttt{\ldots} & \texttt{\ldots} & \texttt{\ldots} & \texttt{\ldots} & \texttt{\ldots}\\
			\bottomrule
		\end{tabular}
         \caption{Data streams for an example of recidivism risk assessment with COMPAS~\cite{COMPAS}.}\label{fig:example1}
	\vspace{-1.5em}
\end{wrapfigure}

As a motivating example, we consider the COMPAS tool developed by Northpointe~\cite{COMPAS}. COMPAS predicts the recidivism risk of defendants in criminal trials in order to assist judges in, e.g., setting bond amounts or in sentencing during trial. Hence, the system gives a prediction on how likely a person is to commit a(nother) crime, and this prediction has a direct impact on criminal sentencing. A retrospective investigation by ProPublica into the predictions by COMPAS during 2013 and 2014 in Broward County, Florida, revealed that the tool is significantly biased against defendants perceived as black~\cite{ProPublica16}. For instance, the false positive rate for black defendants was found to be significantly higher than for white defendants, i.e., black defendants were more likely classified with a high risk of re-offending, without actually committing a crime in the near future. The ultimate vision of our work is that, instead of such a posterior analysis of algorithmic fairness, runtime monitors are deployed that assess the fairness of decision and prediction systems during their execution, in order to raise awareness of unfair treatment early and in this way mitigate unproportional harm put on groups due to an unfair bias. To illustrate our monitoring approach, we will consider a simplified version of the risk assessment setting as shown in Figure~\ref{fig:example1}. This table shows a number of events describing an execution of the COMPAS system that is defined on data streams such as \texttt{event} or \texttt{id}. For example, the first row describes that on the 2nd of January 2013, an individual of group \texttt{A} was screened via COMPAS and assessed to have a high risk of recidivism. A (simplified) algorithmic-fairness specification compares certain conditional probabilities associated with the different groups:
$$
    \big| \, \prob (\texttt{HIGH} \mid \texttt{A}, \texttt{RECIDIVISM}) - \prob (\texttt{HIGH} \mid \texttt{B}, \texttt{RECIDIVISM}) \, \big| \leq \epsilon \enspace .
$$
This condition states that the probability of a re-offending member of group \texttt{A} to be labeled as high-risk is not too far (less than $\epsilon$) from the probability of a re-offending member of group \texttt{B} to be labeled as high-risk. Hence, it compares the \emph{true positive rates} between the two groups.
In this paper, we show how we can use the stream-based monitoring language RTLola to process such data streams and in this way analyze the algorithmic fairness of their underlying system in real-time. The main idea is to automatically partition the stream events into independent trials and to construct RTLola specifications that estimate the conditional probabilities associated with algorithmic-fairness specifications.

\subsection{Outline and Contributions}

The challenges in our stream-based setting are twofold: First, we observe only a single execution of the system but require a larger number of independent trials to reliably estimate the conditional probabilities. Second, the independent trials and also the fairness definitions contain a real-time component. We address the first challenge in Section~\ref{sec:statistics} by defining a principled way to extract individual trials based on a predefined dependence relation between stream events. In Section~\ref{sec:implementation}, we then describe how this can be implemented in the specification language RTLola. We show how we can estimate conditional probabilities over these trials with RTLola, and address the second challenge: RTLola naturally supports reasoning about real-time events, and hence we can use it to collect stream events that are spread throughout time and calculate their relative delay, which allows us to express certain intricacies of algorithmic-fairness specifications, such as an upper time bound between relevant events. We evaluate this RTLola compilation in a case study including both synthetic and real-world benchmarks. For the former, we present a benchmark generator that models application scenarios at a company and a seminar assignment at a university. In both cases, we can easily scale, e.g., the number of applicants, which serves as a stress test for our implementation and allows a thorough comparison with more traditional approaches based on databases. We show that RTLola significantly outperforms database approaches, which suggests that stream-based monitoring is the tool of choice for settings with high data throughput. Moreover, synthetic benchmarks allow us to set a ground truth for the fairness of the decision system, and we show that our monitoring approach can detect unfair systems without raising too many false alarms on fair systems. As a real-world benchmark, we consider the aforementioned recidivism prediction tool COMPAS~\cite{ProPublica16}. Unlike the synthetic benchmarks, this is also an example of a prediction system, such that more complex specifications become relevant. We show that RTLola is able to express these specifications succinctly and effectively alert to unfairness in the prediction system early. All experiments can be found in Section~\ref{Sec:Experiments}.

\paragraph{Contributions.} To summarize, we make the following contributions:

\begin{itemize}
    \item We formalize the estimation of probabilities from single executions in stream-based monitoring.
    \item We implement RTLola monitors that allow the monitoring of a wide range of algorithmic-fairness specifications from the literature.
    \item We present a generator for constructing challenging benchmarks related to algorithmic fairness in job application and university admission.
    \item We perform an extensive experimental evaluation on these synthetic benchmarks, as well as on a real-world data set from the COMPAS tool.
\end{itemize}

\subsection{Related Work}
Efforts of the machine learning community generally aim more at improving the fairness of learned models than rigorously verifying it~\cite{DBLP:journals/csur/MehrabiMSLG21}.
Three categories of mechanisms stand out, namely Pre-Processing~\cite{DBLP:journals/kais/KamiranC11,DBLP:journals/corr/FriedlerSV14}, In-Processing~\cite{DBLP:journals/corr/abs-1803-02453,DBLP:conf/pkdd/KamishimaAAS12}, and Post-Processing~\cite{Pessach2023,Corbett-DaviesP17}.
Our work on monitoring algorithmic fairness is an orthogonal effort that allows us to audit learned systems even when their training process cannot be influenced, as we treat the learned system as a black box.
We present a general approach based on RTLola and encode popular fairness properties, such as \emph{equalized odds}.
These techniques can also be used to encode other fairness properties such as \emph{equal opportunity}~\cite{HardtPNS16} or \emph{counterfactual fairness}~\cite{KusnerLRS17}.

Related to our effort of verifying and testing fairness, a variety of different approaches in the formal methods community exist:
Udeshi et al. \cite{DBLP:conf/kbse/UdeshiAC18} propose an automated and directed testing technique to generate discriminatory inputs for machine learning models.
FairTest~\cite{DBLP:conf/eurosp/TramerAGHHHJL17} is a framework for specifying and testing algorithmic fairness.
A similar approach is given by Bastiani et al. \cite{DBLP:journals/pacmpl/Bastani0S19} by using adaptive concentration inequalities to design a scalable sampling technique for providing fairness guarantees.
Albarghouthi et al. \cite{DBLP:journals/pacmpl/AlbarghouthiDDN17} transform fairness properties as probabilistic program properties and develop an SMT-based technique to verify fairness of decision-making programs. Albarghouthi and Vinitsky~\cite{AlbarghouthiV19} propose a white-box monitoring technique based on adding annotations in a program, but they cannot reason about temporal properties, unlike our approach.
To certify individual fairness, Rouss et al. \cite{DBLP:journals/corr/abs-2002-10312} introduce a local property that coincides with robustness within a particular distance metric.
Another approach is to repair biased decision systems with a program repair technique~\cite{DBLP:conf/cav/AlbarghouthiDD17}. Teuber and Beckert~\cite{TeuberB24} have made an intriguing connection between secure information-flow and algorithmic fairness, and use information-flow tools for verifying fairness of white-box programs.
Henzinger et al.\ propose monitoring of \emph{probabilistic specification expressions (PSEs)}~\cite{HenzingerKKM23a} and extensions~\cite{HenzingerKM23} for monitoring algorithmic fairness properties~\cite{HenzingerKKM23b}. Baum et al.~\cite{BaumBHHLLS24} combine monitoring and input generation for a probabilistic falsification technique aimed at individual fairness. Cano et al.~\cite{Cano} propose fairness shields that combine monitoring and enforcement of fairness properties. In our work, we show that it is possible to use the widely studied formalism of stream-based monitoring languages~\cite{BaumeisterFKS24} to go even further by additionally considering temporal aspects of fairness such as delays between relevant events. Notably, this is possible without any pre-processing of the stream-events that may be needed for, e.g., PSEs, as this is already handled by stream-based monitoring languages. These languages predate the PSE approach by decades~\cite{DBLP:conf/time/DAngeloSSRFSMM05} and have already proven useful in diverse areas such as unmanned aircraft~\cite{cav4,BaumeisterFSST20} and network monitoring~\cite{FaymonvilleFST16}. We use RTLola in this paper, but the general ideas may also be adapted to other stream-based languages, such as TeSSLa~\cite{LeuckerSS0S18} or Striver~\cite{GorostiagaS18}.

The usage of opaque machine-learning models in high-stake scenarios has sparked scholarly debate on its ethics~\cite{BostromYudkowsky14,Matthias04}, as well as extensive governmental regulation~\cite{AIAct,ColoradoAct}. Given that these models promise to be more accurate~\cite{Lin2020} and ultimately even more impartial than human decision makers, there seems to be a clear trend toward further adoption. As we show here, RTLola can be a useful tool for alleviating unintended negative side effects of this trend by promoting effective monitoring of the decision system during deployment.

\section{Preliminaries}

We briefly recall the necessary background on probability theory, algorithmic fairness and stream-based monitoring with RTLola.

\subsection{Probability Theory}

A \emph{probability space} is a tuple $(\Omega,\mathcal{E},\prob)$, where $\Omega$ is a \emph{sample space} and $\mathcal{E}$ is a $\sigma$\emph{-algebra} over $\Omega$, i.e., we have $\emptyset \in \mathcal{E}$, $A \in \mathcal{E} \implies \bar{A} \in \mathcal{E}$, and $A_0,A_1,\ldots \in \mathcal{E} \implies \bigcup_{i=0}^{\infty}A_i \in \mathcal{E}$. Finally, $\prob$ is a \emph{probability measure} $\mathcal{E} \rightarrow \mathbb{R}$, i.e., a non-negative function with $\prob(\Omega) = 1$, $\prob(\emptyset) = 0$ that satisfies countable additivity: For any sequence of pairwise disjoint events $A_0,A_1,\ldots \in \mathcal{E}$ we have that $\prob(\bigcup_{i=0}^{\infty}A_i) = \Sigma_{i=0}^{\infty} \prob(A_i)$. A \emph{random variable} is a function $X : (\Omega,\mathcal{E}) \rightarrow (\Gamma,\mathcal{V})$ that maps elements of the sample space $\Omega$ to some set $\Gamma$ equipped with the $\sigma$-algebra $\mathcal{V}$, such that $X^{-1}(B) \in \mathcal{E}$ for all $B \in \mathcal{V}$. Such an $X$ induces a probability measure on $(\Gamma,\mathcal{V})$ as $\prob(B) = \prob(X^{-1}(B))$ for all $B \in \mathcal{V}$. Lastly, the \emph{conditional probability} of some $A \in \mathcal{E}$ given some $B \in \mathcal{E}$ is defined as $\prob(A \mid B) = \frac{\prob(A \cap B)}{\prob(B)}$.

\subsection{Algorithmic Fairness}\label{prelim:fairness}

Algorithmic fairness is an umbrella term for several specifications that have recently been put forward for decision and classification systems~\cite{Pessach2023}. The general idea is to compare the probabilities of certain good and bad events between social groups, e.g., we may require the probability of a loan request being accepted conditioned on the applicant being in group $A$ to be not too far from the same probability conditioned on the applicant being in group $B$. In the following, we introduce the fairness specifications considered in this work, as they have been proposed in the literature. Note that these pure definitions do not consider timing issues such as the good outcome being obtained within a certain bound. 

The simplest fairness specification is \emph{demographic parity}, which requires the probabilities of good outcomes conditioned on the different groups to differ no more than the predefined parameter $\epsilon$.
\begin{definition}[Demographic Parity~\cite{DworkHPRZ12}] \label{Def: Demographic_Parity}
A decision system for the binary decision $A$ satisfies demographic parity, iff
\begin{align*}\big| \, \prob (A = 1 \mid G = 1) - \prob (A = 1 \mid G = 0) \, \big| \leq \epsilon \enspace . 
\end{align*}
\end{definition}
The value of G represents the group a person belongs to (e.g., male or female), while A indicates the positive outcome. Demographic parity ensures that positive outcomes are assigned to the two groups at a similar rate, but it does not consider background factors that may be relevant to assess the fairness of a system. For instance, men and women may apply unproportionally to different departments of a university, such that the admission process of the university appears to be unfair while the same processes of the individual departments are fair.\footnote{This observation has been termed \emph{Simspon's Paradox} by Colin Blyth~\cite{Blyth72}.} If the existence of such confounding variables is known, it may be more appropriate to use a fairness measure such as \emph{conditional statistical parity}.

 \begin{definition}[Conditional Statistical Parity~\cite{Corbett-DaviesP17}] \label{Def:Conditional_Parity} A decision system for the binary decision $A$ satisfies conditional statistical parity, iff
\begin{align*}
    \big| \, \prob (A = 1 \mid L = 1,G = 1) - \prob (A = 1 \mid L = 1,G = 0) \, \big| \leq \epsilon \enspace .
\end{align*}
\end{definition}
Conditional statistical parity states, similar to demographic parity, that people from different groups should have an equal probability of positive outcomes. Additionally, it further conditions the probability on other legitimate factors $L$, e.g., confounding variables such as the department that students apply to. These factors have to be determined a priori and are based on the background knowledge of the specifier.

While the above parity measures define a notion of fairness for decision systems with binary outcomes, many fairness issues arise also for prediction systems that, e.g. classify the recidivism risk of defendants in criminal trials~\cite{ProPublica16}. Fairness of such systems is more accurately described by comparing the true and false positive rates between groups, as done by the \emph{equalized-odds} fairness measure.

\begin{definition}[Equalized Odds~\cite{HardtPNS16}]\label{def:equalodds} \label{Def:Equalized_Odds} A prediction system $\hat{Y}$ for the outcome $Y$ satisfies equalized odds, iff
\begin{align*}
    &\big| \, \prob (\hat{Y} = 1 \mid G = 1, Y = 0) - \prob (\hat{Y} = 1 \mid G = 0, Y = 0) \, \big| \leq \epsilon \enspace \text{and}\\
    &\big| \, \prob (\hat{Y} = 1 \mid G = 1, Y = 1) - \prob (\hat{Y} = 1 \mid G = 0, Y = 1) \, \big| \leq \epsilon \enspace .
\end{align*}
\end{definition}
Here, $\hat{Y}$ describes the predicted value, while $Y$ is the true value of an outcome to be predicted. Hence, the equalized-odds measure requires the differences between the false positive rates (FPR) and the differences between the true positive rates (TPR) of all pairs of groups to be within a predefined bound $\epsilon$.

\subsection{Stream-based Monitoring with \rtlola}
\label{sec:prelim:rtlola}
In this work, we use the stream-based specification language \rtlola to monitor the previously described fairness definitions.
\rtlola uses stream-equations to translate streams of input data to output streams and trigger conditions that describe violations of the specification.
We illustrate the \rtlola language with a small example and refer for more details to \cite{BaumeisterFKS24,cav4,DBLP:conf/atva/FinkbeinerKS23}.
\begin{example}[RTLola Example]
\label{ex:lola}
\begin{lstlisting}
	input user_id : UInt64, value : Int64
	output amount(user)
		spawn with user_id
		eval when user_id = user with value + amount(user).last(or: 0)
	trigger @value@ amount.aggregate(over_instances: all, using: max).defaults(to: 0) > 500 "Upper Limit Violation"
	trigger @1Hz@ value.aggregate(over: 1s, using: count) > 5 "Too many transactions"
\end{lstlisting}
The specification declares two input streams describing a transaction to a user: The input stream \lstinline!user_id! encodes a unique identifier for each user and \lstinline!value! represents the amount.
Next, the output stream \lstinline!amount! sums up the values per user using parameterization.
With parameterization, the output stream describes a set of instances and the specification can refer to each instance with the parameter, in this example the parameter \lstinline!user!.
The spawn declaration describes when a new instance is added to this set, in our case for every new \lstinline!user_id!. The eval declaration describes for each instance when a new value is computed with the \lstinline!when!-clause and the computation of this value with the \lstinline!with!-clause.
Here, each instance of the \lstinline!amount! stream is computed when the \lstinline!user_id! is equal to the instance parameter and the new value is computed as the sum of the previous value of this instance and the current value of the \lstinline!value!-stream.
The first trigger then aggregates over all instances of the \lstinline!amount! stream, takes the maximum value, and compares this value against a threshold.
Since in theory this access could fail, we need to provide a default value.
If this condition is true, the generated monitor for this specification emits the corresponding trigger message.
The second trigger checks the number of transactions over the last five seconds, illustrating the real-time capabilities of \rtlola.
\end{example}

The semantics of \rtlola is defined over a collection of timed data streams and intuitively checks whether the values in the collection correspond to the computed values for the stream equation.
Additionally, it validates that the time is monotone.
\begin{definition}[Data Streams]\label{def:streams}
A collection of timed data streams $\world \in \World$ over a set of input streams $\srin$ and output streams $\srout$ is the combination of a $\StreamMap$ and a $\stimeMap$. 
\[
\begin{array}{r l}
	\Stream &:= \instanceId \rightarrow \stime \rightarrow \svalue\\
	\StreamMap &:= \sr \rightarrow \Stream\\
	\stimeMap & := \stime \rightarrow \mathbb{R}\\
	\World & := \StreamMap \times \stimeMap
\end{array}
\]
\end{definition}
\Cref{fig:data_streams} gives an intuition on the data stream representation based on the specification in \Cref{ex:lola}.
The $\stimeMap$ is a total function from the discrete timestamps, indicated at the top of the figure, to a real-time value.
In the examples, the first three events arrive at the timestamps 0.6, 0.8, and 2.4.
Given $\world = (\mathit{streams}, \mathit{times}) \in \World$, we use $\world(t) := \mathit{times}(t)$ to get the real-time value of a discrete timestamp $t \in \stime$.
The $\StreamMap$ assigns each stream identifier and instance to an infinite sequence of optional values, where $\bot$ indicates that the stream instance does not produce a value.
In our example, $\world(user\_id)(\top)$ represents the infinite sequence of the input stream, and $\world(user\_id)(\top)(1)$ returns the value of the input stream at time 1.
Note that we use $\top$ as the instance identifier if the stream is not parameterized, i.e., only one stream instance exists in the $\StreamMap$.
In contrast, the output stream \lstinline{amount} is parameterized, such that different instances (e.g., $\world(amount)(1)$) exist.
Formally, infinite sequences are represented by total functions, and we define the access functions $\world(\sid) := \mathit{streams}(\sid)$ for the stream $\sid \in \sr$, $\world(\sid)(i) := \mathit{streams}(\sid)(i)$ to access stream instances $i \in \instanceId$ of stream $\sid$, and $\world(\sid)(i)(t)$ for the stream instance value at discrete timestamp $t$.
\begin{figure}[t]
\resizebox{\linewidth}{!}{
\begin{tikzpicture}[node distance = 0.7cm and 2cm]
	\def\xsep{4.5}
	\def\yskip{5.15}
	\node (timelabel) {\emph{Time}};
	\node[below=of timelabel.west,anchor=west] (timemaplabel) {\emph{TimeMap}};
	\node[below=0.8cm of timemaplabel.west,anchor=west] (streammaplabel) {\emph{StreamMap}};

	\node[below=of streammaplabel.west,anchor=west,xshift=3mm] (useridlabel) {Stream $\world(user\_id)$};
	\node[below=of useridlabel.west,anchor=west,xshift=3mm] (useridlabel1) {$\world(user\_id)(\top)$};
	\node[below=0.9cm of useridlabel1.west,anchor=west,xshift=-3mm] (amountlabel) {Stream $\world(amount)$};
	\node[below=of amountlabel.west,anchor=west,xshift=3mm] (amountlabel1) {$\world(amount)(0)$};
	\node[below=of amountlabel1.west,anchor=west] (amountlabel2) {$\world(amount)(1)$};
	\node[below=of amountlabel2.west,anchor=west] (amountlabel3) {$\world(amount)(2)$};

	\foreach \i in {0,1,2} {
		\node at (\i*\xsep+\yskip, |- timelabel) {\i};
	}
	\node at (2.65*\xsep+\yskip, |- timelabel) {$\cdots$};

	\foreach \i/\label in {
        0/$\world(0) = 0.6$,
        1/$\world(1) = 0.8$,
        2/$\world(2) = 2.4$
    } {
		\node[draw,minimum width=\xsep cm] at (\i*\xsep+\yskip, |- timemaplabel) (timemapbox) {\label};
	}
	\node at (2.65*\xsep+\yskip, |- timemaplabel) {$\cdots$};

	\foreach \i/\label in {
		0/$\world(user\_id)(\top)(0) = 2$,
		1/$\world(user\_id)(\top)(1) = 0$,
		2/$\world(user\_id)(\top)(2) = 2$
	} {
		\node[draw,minimum width=\xsep cm] at (\i*\xsep+\yskip, |- useridlabel1) (useridbox) {\label};
	}
	\node at (2.65*\xsep+\yskip, |- useridlabel1) {$\cdots$};

        \foreach \i/\label in {
    			0/$\world(amount)(0)(0) = \bot$,
    			1/$\world(amount)(0)(1) = 2$,
    			2/$\world(amount)(0)(2) = \bot$
    		} {
    			\node[draw,minimum width=\xsep cm] at (\i*\xsep+\yskip, |- amountlabel1) (amount1box) {\label};
    		}
    	\node at (2.65*\xsep+\yskip, |- amountlabel1) {$\cdots$};
        \foreach \i/\label in {
			    0/$\world(amount)(1)(0) = \bot$,
			    1/$\world(amount)(1)(1) = \bot$,
			    2/$\world(amount)(1)(2) = \bot$
		} {
			\node[draw,minimum width=\xsep cm] at (\i*\xsep+\yskip, |- amountlabel2) (amount2box) {\label};
		}
		\node at (2.65*\xsep+\yskip, |- amountlabel2) {$\cdots$};
		\foreach \i/\label in {
			0/$\world(amount)(2)(0) = 3$,
			1/$\world(amount)(2)(1) = \bot$,
			2/$\world(amount)(2)(2) = 5$
		} {
			\node[draw,minimum width=\xsep cm] at (\i*\xsep+\yskip, |- amountlabel3) (amount3box) {\label};
		}
		\node at (2.65*\xsep+\yskip, |- amountlabel3) {$\cdots$};

    \foreach \i in {timemapbox,useridbox,amount1box,amount2box,amount3box} {
        \draw ([yshift=-0.05mm]\i.north -| 2.5*\xsep+\yskip,) -- ++(1cm,0);
        \draw ([yshift=0.05mm]\i.south -| 2.5*\xsep+\yskip,) -- ++(1cm,0);
    }

	\begin{scope}[on background layer]
	\fill[red!20!white] ([yshift=1mm]timemaplabel.north west) rectangle ([yshift=-1mm]2.85*\xsep+\yskip, |- timemaplabel.south);
	\fill[cyan!20!white] (streammaplabel.north west) rectangle ([yshift=-3mm]2.85*\xsep+\yskip, |- amountlabel3.south);
	\fill[cyan!30!white!90!black] (useridlabel.north west) rectangle ([yshift=-1mm]2.78*\xsep+\yskip, |- useridlabel1.south);
	\fill[cyan!30!white!90!black] (amountlabel.north west) rectangle ([yshift=-1mm]2.78*\xsep+\yskip, |- amountlabel3.south);
	\end{scope}
\end{tikzpicture}}

\caption{
    The data streams exemplified on the specification from \Cref{ex:lola}.
}
\label{fig:data_streams}
\end{figure}

The set of \emph{stream events} of $\world$ is defined as  $\mathit{Events}(\world) := \{(r,f) \mid \forall \sid \in \sr , i \in \instanceId . \, \world(\sid)(i)(\world^{-1}(r)) = f(\sid)(i)\} \subseteq \mathbb{E}_\world := \mathbb{R} \times (\sr \rightarrow \instanceId \rightarrow \svalue)$. Hence, $\mathbb{E}_\world$ denotes the set of all conceivable stream events over the datatypes defined by $\world$, while $\mathit{Events}(\world)$ denotes the concrete events appearing in $\world$.
For our example above, $\mathit{Events}(\world)$ would map each real-time timestamp to the corresponding column in the figure.

\section{Statistical Estimates from Data Streams}\label{sec:statistics}

In this section, we outline the formal background for our RTLola specifications that estimate algorithmic fairness properties. We first describe how to extract multiple samples from a single execution of our system, we then describe how to use random variables to describe fairness properties in this setting, and lastly how we estimate the probability of events over these random variables.

\subsection{Extracting Independent Trials from Data Streams}
\label{sec:statistics:independent-trials}
The central challenge in our setting is that we observe only a single execution of the system under scrutiny but want to perform a statistical estimation that naturally gets more accurate the more samples become available. We utilize the fact that in our applications, the single system execution describes a number of independent trials pertaining to the specification we care about, e.g., a single execution of the COMPAS tool for assessing the recidivism risk of defendants describes a large number of independent risk screenings. Hence, we propose a principled way to extract multiple samples from the observed system execution. At its core lies the definition of the probability space $(\Omega_\world,\mathcal{E}_\world,\prob_\world)$ associated with the data streams $\omega \in \World$. The sample space $\Omega_\world$ is constructed as the set of all possible sequences of dependent events, which we identify through a \emph{dependence relation} $\delta \subseteq \mathbb{E}_\world^2$. This predefined $\delta$ is an equivalence relation over the stream events $\mathbb{E}_\world$ whose equivalence classes define the possible sets of events that form mutually independent trials. Elements of the sample space $\Omega_\world$ are ordered subsets of such dependent events: $\Omega_\world := \{E_0 \ldots E_n \in E^n \mid \forall \, 0 \leq i \leq j \leq n . \, t(E_i) \leq t(E_j) \land \delta(E_i,E_j) \}$ and we take $\mathcal{E}_\world$ simply as the powerset of $\Omega_\world$, while $\prob_\world$ is unknown to us.

\begin{example}
    Consider the COMPAS recidivism risk assessment tool described in Section~\ref{sec:motivation} and the corresponding data streams illustrated in Figure~\ref{fig:example1}. We assume that the outcomes of individual screenings do not affect each other, and hence define the dependence relation such that two events are dependent if they refer to the same defendant (identified through the stream \texttt{id}), i.e., $\delta := \{ (E_0,E_1) \mid E_0(\text{\texttt{id}}) = E_1(\text{\texttt{id}}) \}$. Consequently, the data streams $\world$ illustrated in Figure~\ref{fig:example1} describe the following samples $s_{0,1,2} \in \Omega_\world$. 
    \begin{align*}
        &s_0 = (0.0,\texttt{SCREEN},\texttt{0},\texttt{A},\texttt{HIGH})\ldots\\
        &s_1 = (0.0,\texttt{SCREEN},\texttt{1},\texttt{B},\texttt{LOW})(1.0,\texttt{RECIDIVISM},\texttt{1},\texttt{-},\texttt{-})\ldots\\
        &s_2 = (1.0,\texttt{SCREEN},\texttt{2},\texttt{B},\texttt{HIGH})\ldots
    \end{align*}
    Hence, our dependence relation $\delta$ partitions the data streams of the system into independent sequences of stream events, that naturally grow the more events are produced by the system. Note that the first components in the stream events with the values $0.0$ and $1.0$ encode the dates, i.e., $\texttt{2013-01-02}$ and $\texttt{2013-01-03}$, via the $\StreamMap$ as outlined in Definition~\ref{def:streams}.
\end{example}

\subsection{Defining Indicator Variables}
\label{sec:statistics:indicator-variables}
Having defined our probability space through a dependence relation $\delta$, the next step is to define Bernoulli random variables $X : \Omega_\world \rightarrow \{0,1\}$ that serve as indicator variables for the events relevant to algorithmic fairness.

\begin{example}
\label{ex:indicatior-variables}
    For instance, we may want to specify equalized odds (Definition~\ref{def:equalodds}) for the COMPAS risk assessment tool from Section~\ref{sec:motivation}. We may naturally define the prediction $\hat{Y}$ for a defendant associated with the sample $\omega$ as $\hat{Y}(\omega) := \exists i . \, \omega(\texttt{event})(i) = \texttt{SCREEN} \land \omega(\texttt{risk})(i) = \texttt{HIGH}$, and similarly, the true outcome is defined as $Y(\omega) := \exists i . \, \omega(\texttt{event})(i) = \texttt{RECIDIVISM}$. Here, we quantify over the time stamps $i$. Membership to, e.g., group \texttt{A} is captured by $G_{\texttt{A}}(\omega) := \exists i . \, \omega(\texttt{event})(i) = \texttt{SCREEN} \land \omega(\texttt{group})(i) = \texttt{A}$. It is also possible to define a sanity check as an additional variable that we condition on.
    For example, we may only consider recidivism events that happen less than two years after a screening event, as this is the specific time horizon that the COMPAS tool is targeting~\cite{ProPublica16,COMPAS}.
    We can achieve this by utilizing the real-time information of the stream events with the variable $Y_{<2y} := \exists i,j.\,  \omega(\texttt{event})(i) = \texttt{SCREEN} \land \omega(\texttt{event})(j) = \texttt{RECIDIVISM} \land \world(j) - \world(i) < 730.0$. Hence the FPR part of a specification of equalized odds with $\epsilon = 0.1$ is:
    \begin{align*}
        \varphi := \big| \, &\prob (\hat{Y} = 1 \mid G_{\texttt{A}} = 1, Y_{<2y} = 1) \, - \prob (\hat{Y} = 1 \mid G_{\texttt{A}} = 0, Y_{<2y} = 1) \, \big| \leq 0.1 \enspace .
    \end{align*}
\end{example}

\subsection{Maximum A Posteriori Estimation} \label{sec:MAP}

Since during monitoring we obtain samples sequentially, the first samples have an unproportionally large impact on the assessment of fairness at the start of monitoring, since the estimation of the conditional probabilities in a formula like $\varphi$ only gets more robust over time. Hence, we use methods from Bayesian statistics to control the trigger behavior of the monitor at the start of an execution: \emph{maximum a posteriori (MAP)} estimation~\cite{MAPestimation} allows us to take a prior belief about the conditional probabilities that make up the fairness specifications into consideration, as well as a degree of confidence therein. Formally, for every conditional probability $\Theta = \prob(A \mid B)$ in our specification we require a prior $\gamma$ and a confidence $\kappa$. Then, the estimate $\hat{\Theta}$ is given as:
\begin{align*}
  \hat{\Theta} = \frac{S_{A \cap B} + \gamma(\omega) \kappa}{S_B + \kappa} \enspace ,
\end{align*}
where $S_{A \cap B}$ is the number of samples that satisfy $A$ and $B$, while $S_{B}$ is the number of samples satisfying $B$. The parameters $\gamma$, $\kappa$ and $\epsilon$ suffice to achieve sufficient initial robustness of the monitor, which we demonstrate experimentally in Section~\ref{Sec:Experiments}. The longer the observed system execution gets and the more samples become available, the less influence these parameters have on the monitor verdict.

\paragraph{Dynamic Updating of the Prior Belief.} While MAP is a standard method from statistics, we face unique challenges when dynamically analyzing data streams, since we only have limited knowledge about the monitored system. Certain background knowledge like how many free places and applicants emerge during the execution may change the prior belief we have about the conditional probabilities. For instance, we may know that a university always fills all seminars with students, but the chance of an individual student's application to be accepted of course still depends on the number of seminar places and the number of other students applying. To account for such dynamic updates to the prior belief, we consider the prior $\gamma$ to be a function of the data streams $\omega$, such that it may be defined, e.g., as the ratio of places and applying students. 

\section{Implementation in \rtlola}\label{sec:implementation}

This section describes the implementation of the fairness definitions from \Cref{prelim:fairness} in the stream-based specification language \rtlola.
In general, each fairness specification follows the same structure:
First, we extract information on independent trials from the input data and store it in parameterized streams that directly correspond to the indicator variables that are relevant in a given fairness specification. These variables can use the full power of \rtlola expressions such as  stream aggregations and real-time properties. We then build accumulators that are used in estimating the conditional probabilities. Last, we define trigger conditions that indicate that the estimates violate the fairness specification.

\subsection{Implementation of Equalized Odds for the COMPAS Tool}
We illustrate this principle by discussing the implementation of equalized odds (cf.~\Cref{ex:indicatior-variables}). The RTLola specification for this fairness property, in the context of the COMPAS system, is shown in \Cref{fig:rtlola:compas}.
The specification is defined over input data streams that encode the relevant events of the COMPAS system as described in \Cref{sec:motivation}:
The \lstinline!"SCREEN"! event includes the unique identifier of a defendant in the input stream \lstinline!id!, their group attribute in the input stream \lstinline!group! and the COMPAS score describing the predicted likelihood of that person re-offending in the input stream \lstinline!score!. The COMPAS score is an integer value between $0$ and $10$ as in the original data set. We use the same classification of any score above $6$ as high risk as used by ProPublica~\cite{ProPublica16} in the original investigation.
If the defendant re-offends, the second event \lstinline!"RECIDIVISM"! is given to the monitor together with the identifier of the defendant. The timestamps of these events are implicitly included through RTLola.

\begin{figure}
	\centering
	\begin{lstlisting}
input event : String
input id : Int64
input group : String
input score : Int64

/// Defendant Information
output days_per(i)
  spawn with id
  eval @Global(1d)@ with days_per(i).last(or:  0) + 1
  close when days_per(i) = 730
output has_re(i)
    spawn with id
    eval when id == i with event == "RECIDIVISM"
    close @Global(1d)@ when days_per(i).hold(or: 0) = 730
output tp_event(i, g, s)
  spawn with (id, group, score)
  eval @Global(1d)@
  	when days_per(i).hold(or: 0) = 730 $\wedge$ has_re(i).aggregate(over: 730d, using: $\exists$) with s > 6(*\label{ln:indicator_y}*)
  close @Global(1d)@ when days_per(i).hold(or: 0) = 730

/// TP Ratio
output abs_re(g) : UInt64(*\label{ln:abs_re}*)
    spawn with group
    eval @Global(1d)@ with abs_re(g).last(or:  100) +
    tp_event.aggregate(over_instances: All(ii, ig, is => ig = g), using: count)
output abs_hr_re(g) : UInt64(*\label{ln:abs_hr_re}*)
  spawn with group
  eval @Global(1d)@ with abs_hr_re(g).last(or:  50) +
  	tp_event.aggregate(over_instances: All(ii, ig, is => ig = g), using: sum)
output tp_ratio(g)
    spawn with group
    eval when abs_re(g) != 0
    	with cast<UInt64, Float64>(abs_hr_re(g)) / cast<UInt64, Float64>(abs_re(g))(*\label{ln:tp_ratio_end}*)
    	
/// Equalized Odds: True Positive
trigger @1d@ tp_ratio.aggregate(over_instances: all, using: max).defaults(to: 0.0) - tp_ratio.aggregate(over_instances: all, using: min).defaults(to: 0.0) > 0.1(*\label{ln:trigger}*)
	\end{lstlisting}
	\caption{RTLola specification computing and checking the differences of the true positive ratios between all groups, which makes up one half of the equalized-odds specification for the COMPAS data set.}
	\label{fig:rtlola:compas}
\end{figure}

\paragraph{Storing Independent Trials in Parameterized Streams.}

The specification uses three parameterized output streams to store the relevant information of independent trials, where the parameter  \lstinline!i! identifies the trial, e.g., an individual defendant. The streams are \lstinline!days_per!, \lstinline!has_re! and \lstinline!tp_event!. Each of these streams  has a lifecycle of exactly 730 days after screening the defendant.
In \rtlola, this lifecycle is represented with the \lstinline!spawn!, starting the lifecycle with the first occurrence for each identifier, and the \lstinline!close! declaration, ending the lifecycle when the associated condition is satisfied. The output stream \lstinline!days_per! counts the number of days after the screening of the defendant and the output stream \lstinline!has_re! maps a \lstinline!"RECIDIVISM"! event to the defendant.
Then, the output stream \lstinline!tp_event! synchronizes all information about one defendant after 730 days. This realizes the extraction of independent trials as described formally in \Cref{sec:statistics:independent-trials}. For example, the indicator variable $Y_{<2y}$ is described with the second clause of the eval-when declaration of the stream using stream aggregation (line~\ref{ln:indicator_y}), i.e., \lstinline!has_re(i).aggregate(over: 730d, using: $\exists$)!.
This expression checks if the defendant re-offended during a timeframe of 730 days using stream aggregation and follows the definition from \Cref{ex:indicatior-variables}. The stream \lstinline!tp_event! is additionally parametrized with the group and the score of the defendant from which we can derive the indicator variables $G_{\texttt{A}}$ and $C$ directly. After the indicator variables are computed and used by the accumulators as described in the following paragraph, we close these stream instances to free the underlying memory since their value is not required after the first use of the variable.

\paragraph{Accumulator Variables and MAP Estimation.}

The specification then stores the accumulated information for each group using stream parameterization, where this time the parameter \lstinline!g! identifies the group associated with the stream. It uses the stream \lstinline!abs_re! to count the number of defendants that re-offendend in a given group (line~\ref{ln:abs_re}). Similarly, the stream \lstinline!abs_hr_re! counts the number of re-offenders per group that were scored as high-risk by COMPAS  (line~\ref{ln:abs_hr_re}). The parameterized stream \lstinline!tp_ratio! then computes for each group the true-positive ratio $\prob (\hat{Y} = 1 \mid G = g, Y = 1)$, i.e., the probability that a person was assigned a high-risk score under the condition that this person has re-offended. To encode the MAP estimation from \Cref{sec:MAP}, we assign the \lstinline!abs_re! and \lstinline!abs_hr_re! streams different default values when accessing the previous value, which effectively initializes the streams with these default values at the first time point.
Finally, the trigger (line~\ref{ln:trigger}) encodes a violation of the fairness definition using the following underlying formula:
\begin{align*}
    &\mathit{max}_{g \in G}\{\prob (\hat{Y} = 1 \mid G = g, Y = 1) \} - \mathit{min}_{g \in G}\{\prob (\hat{Y} = 1 \mid G = g, Y = 1) \} \leq \epsilon.
\end{align*} Here, $G$ is the set of all groups. Hence, this formula takes the maximum difference between \emph{any} two groups and compares it against the threshold $\epsilon$. This suffices to infer a violation in all cases. Additionally, the exact values of the ratios can be read from the parameterized streams such as \lstinline!tp_ratio!. The full specification for equalized odds extends this principle to the false positive ratio by defining parameterized streams \lstinline!abs_not_re! to count the defendants per group that did not re-offend, \lstinline!abs_hr_not_re! to count the number of these that were screened high-risk, and \lstinline!fp_ratio! for the resulting ratio. Additionally, the trigger condition is extended to account for all pairs of parameters of the \lstinline!fp_ratio! stream. The experimental results of running this specification on the COMPAS data from the original ProPublica investigation can be found in \Cref{sec:compas_eval}.

\section{Case Studies}\label{Sec:Experiments}

We specified all algorithmic fairness requirements defined in Section~\ref{prelim:fairness} with RTLola in a similar way as outlined for equalized odds in Section~\ref{sec:implementation}. In this section, we report on experiments with these fairness specifications in a variety of settings\footnote{Our artifact is available on Zenodo: \url{https://doi.org/10.5281/zenodo.14627198}.}. We first consider synthetically constructed data streams related to hiring and application scenarios that allow us to study the utility and efficiency of the approach under varying assumptions. Afterward, we consider data from the COMPAS recidivism risk assessment tool discussed in Section~\ref{sec:motivation} to assess the utility of our tool in a real-world setting. The experiments were conducted with Ubuntu 24.04, a 4-core Intel i5 2.30GHz processor, as well as 8GB of memory.

\subsection{Synthetic Scenarios}

Our two synthetic scenarios deal with hiring done by a company and seminar assignments at a university. For the hiring scenario, we make the simplifying assumption that the company has no fixed limit on the number of employees it can hire. For the seminar assignment, we assume that each seminar has a fixed number of places. Both scenarios are synthesized from a generator script that allows us to specify and scale a number of interesting parameters such as the number of applicants and seminars, as well as the number of places per seminar. The input streams of both scenarios encode the individual applicants and events related to them, i.e., there is an event for an applicant with a specific \textit{id}, \textit{gender}, and \textit{qualification}. For the seminar assignment system, also an input \textit{seminar} to indicate which seminar the applicant applies to, such that we can also monitor conditional statistical parity in addition to demographic parity. Further, seminars have a predefined maximum number of places.
Additionally, there is a separate input stream \textit{accepted} that gives the IDs of accepted applicants.
The generator allows to specify which decision algorithm should be used. We discuss these with the experimental results in the following.

\begin{figure}[t]
\centering
\begin{subfigure}{0.48\linewidth}
		\centering
\includegraphics[scale=.37]{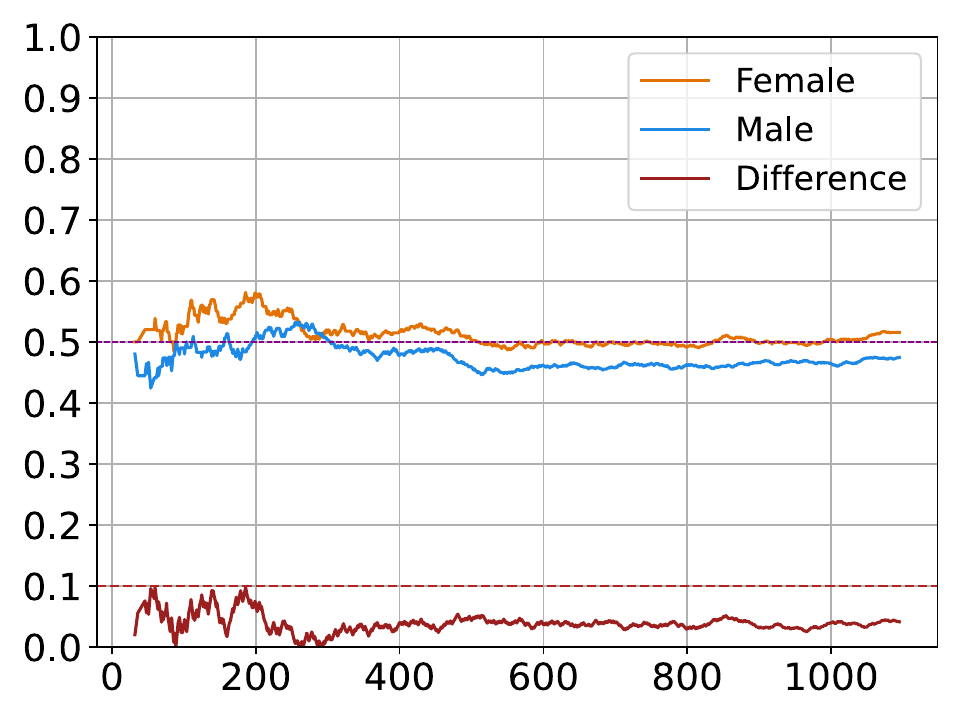}
\caption{Fair algorithm.}
\end{subfigure}\hspace{6pt}\begin{subfigure}{0.48\linewidth}
		\centering
\includegraphics[scale=.37]{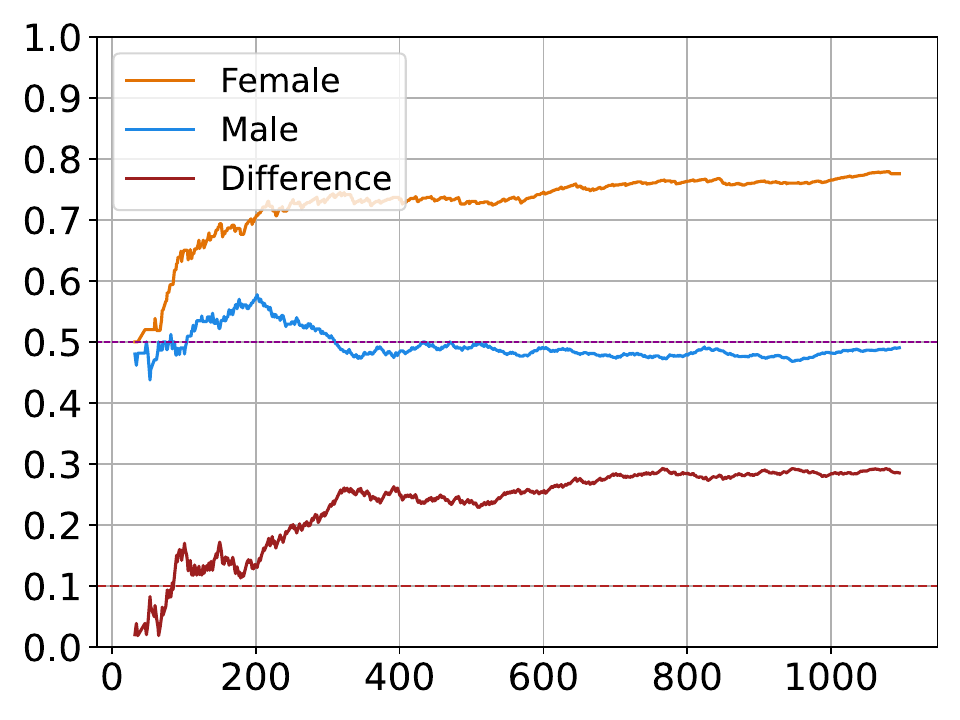}
\caption{Unfair algorithm.}
\end{subfigure}
\caption{Demographic parity of hiring algorithms. The Female and Male lines correspond to the estimates $\hat{\Theta}_{F,M}$ for women and men, respectively. The Difference line shows the absolute difference between these ratios, while the dashed red line at $0.1$ indicates the threshold parameter $\epsilon$ (cf.\ Definition~\ref{Def: Demographic_Parity}).}
\label{fig:DemoParity_Fair}
\end{figure}

\paragraph{Company Hiring.} This scenario modeling a hiring system at a company consists of truly independent trials. We can adjust the probabilities $\Theta_{F} = \prob(A = 1 \mid F = 1)$ and $\Theta_{M} = \prob(A = 1 \mid M = 1)$ with which women and men get accepted, respectively. In this way, we can compare the monitoring outcome of a hiring system that is unfair by construction to a truly fair one. In the unfair system, we set the probability to be accepted for men to $\Theta_{M} = 0.5$ and for women to $\Theta_{F} =0.2$, while both are $0.5$ in the fair system. We then monitor for demographic parity (cf.\ Definition~\ref{Def: Demographic_Parity}). The prior (cf.\ Section \ref{sec:MAP}) is set to $0.5$ with a confidence of $24$ (a more detailed discussion on how to set these parameters follows in the next paragraph). Graphs for the estimated conditional probabilities $\hat{\Theta}_{F,M}$, their difference, and the trigger condition (based on $\epsilon = 0.1$) are depicted in Figure~\ref{fig:DemoParity_Fair}. As we can see, the fair algorithm stabilizes far below the trigger threshold. The diffuse behavior at the start, which usually would result in a number of false alarms, is held back by our MAP approach, such that no triggers are thrown. In contrast, after around time point $85$, the unfair algorithm constantly raises triggers indicating unfairness, as the difference of the conditional probabilities stabilizes far above the threshold of $0.1$, overpowering the prior belief. These results confirm that monitoring can adequately discern between unfair and fair systems after a reasonably small number of decisions has been made.

\paragraph{University Application.} 

\begin{wrapfigure}{r}{0.58\linewidth}
\centering
\vspace{0.25em}
\includegraphics[scale=0.34]{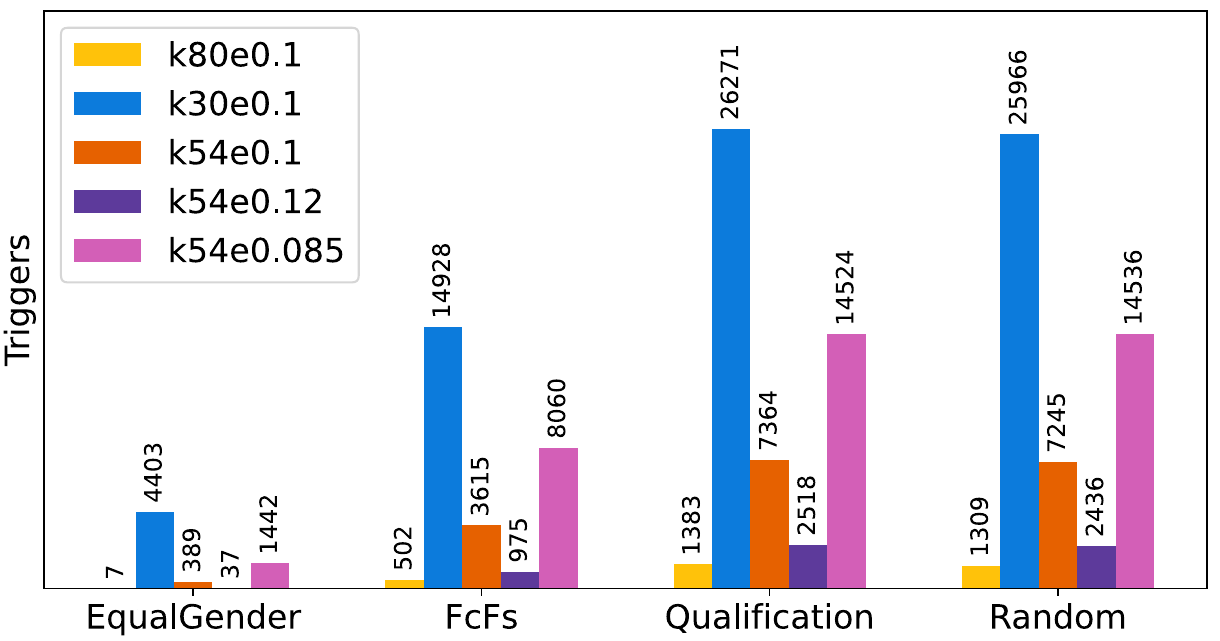}
\caption{Number of triggers thrown for different values of confidence $\kappa$ (k) and threshold $\epsilon$ (e).}
\label{fig:DemoParity_Trigger_Confidence}
\vspace{-1.5em}
\end{wrapfigure}

How to choose the right parameters? We now show that synthetic experiments can be an effective way to choose the confidence $\kappa$ and threshold $\epsilon$. We consider different decision-making algorithms for distributing places to applicants. This lets us explore how the different parameters influence the number of triggers on different algorithms. 
The first algorithm is \emph{First Come First Served (FCFS)}, which accepts the first people applying regardless of other attributes.  The second is \emph{Randomize}, which picks randomly in the pool of applicants for a given seminar. The third algorithm, called \emph{Qualification}, picks the most qualified people for each seminar. The last algorithm is \emph{EqualGender}, which tries to ensure the same acceptance rates for all groups in the long run. Note that demographic parity does not take into account additional attributes such as the qualification, and hence the fairness of, e.g., the Qualification algorithm completely depends on the randomization of the qualification values. Similarly, the fairness of FCFS depends on the application times which are generated randomly. 
In Figure~\ref{fig:DemoParity_Trigger_Confidence}, we compare the number of thrown triggers for the different parameters on two thousand generated scenarios for every algorithm with a hundred applicants each. Our specification of demographic parity with parameters $\kappa = 54$ and $\epsilon = 0.085$ gets violated 1031 times over all the scenarios generated with the EqualGender algorithm (which have 200000 distinct events).
A general trend that can be inferred from Figure~\ref{fig:DemoParity_Trigger_Confidence} is that parameter values that are too low lead to a large amount of triggers. Finding the right parameters requires estimating how many applicants are expected, and selecting them to achieve a desired contrast between the different algorithms on the simulated scenarios. For instance, a confidence of $80$ and threshold of $0.1$ results in $187\times$ as many triggers on the random algorithm than on the EqualGender algorithm, while a confidence of $30$ and threshold of $0.1$ results only in around $6\times$ as many triggers on the random algorithm.

\begin{figure}[t]
\centering
\begin{subfigure}{0.48\linewidth}
\includegraphics[scale=.264]{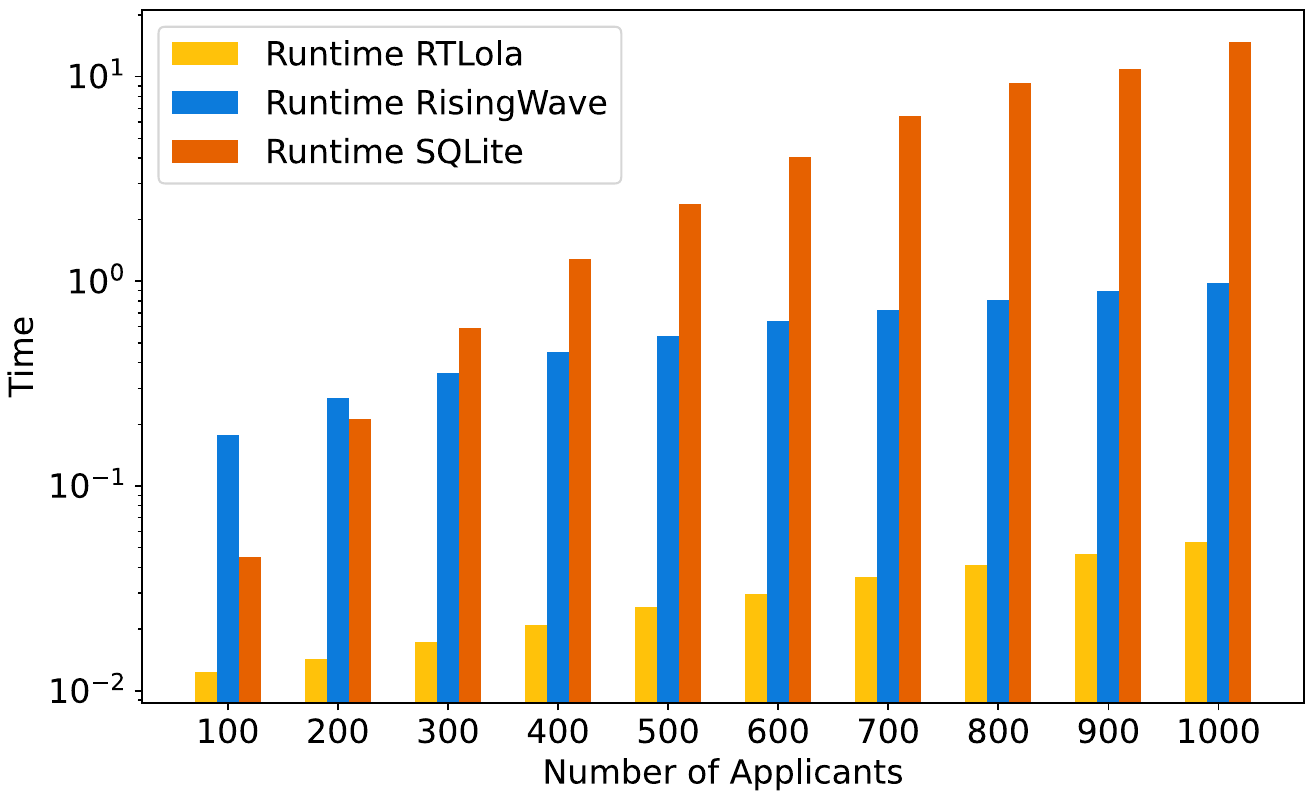}
\caption{Demographic parity.}
\end{subfigure}\hspace{8pt}\begin{subfigure}{0.48\linewidth}
		\centering
\includegraphics[scale=.264]{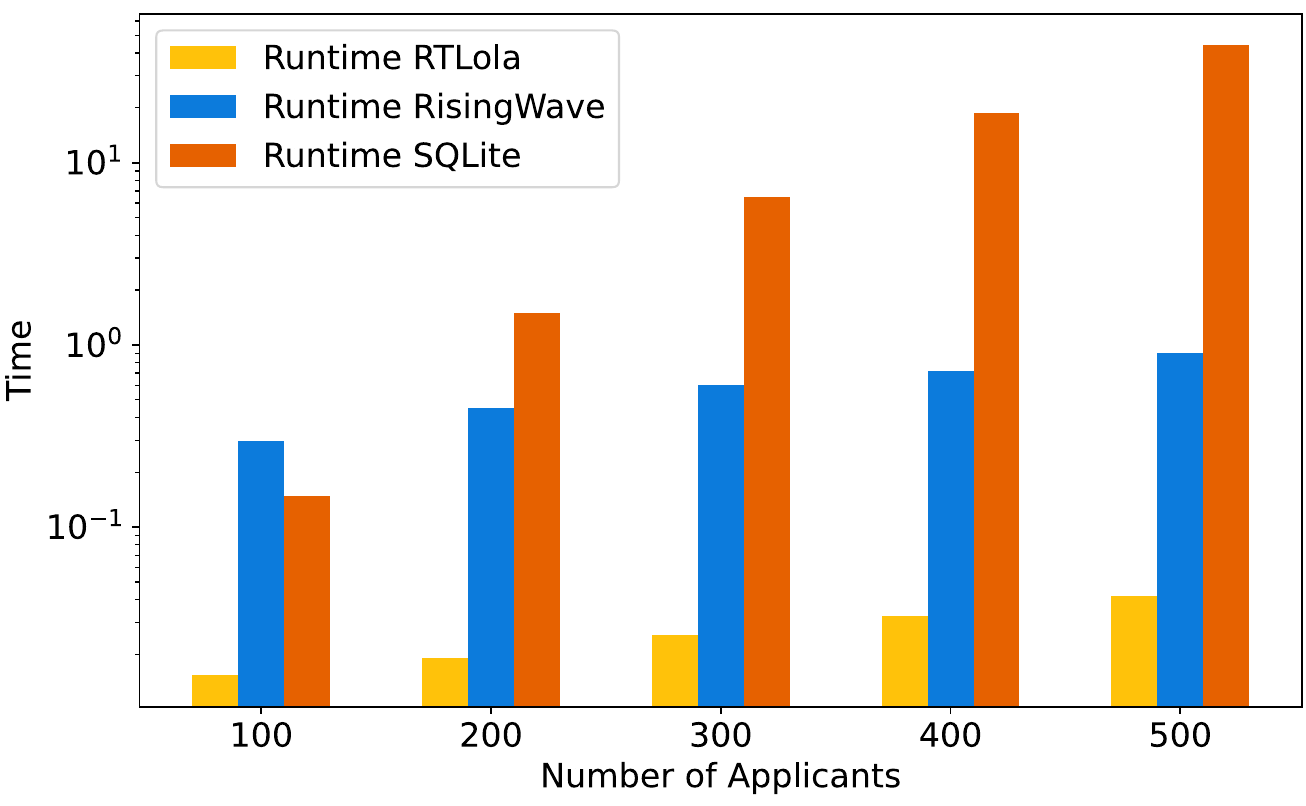}
\caption{Conditional statistical parity.}
\end{subfigure}
\caption{Runtime comparison between monitoring and database implementations. The bars report the average runtime over ten generated scenarios.}
\label{fig:DemoParity_Runtime}
\end{figure}

\paragraph{Runtime Comparison.} It is a viable question to ask what advantages monitoring with a stream-based specification language has over a simple database implementation. Therefore, we have compared our approach with a naïve implementation using SQLite, and an advanced implementation using RisingWave~\cite{WangL22}, a state-of-the-art streaming database~\cite{FragkoulisCKK24}. For the databases we first defined a SQL query that encodes the fairness specifications and returns a Boolean value, similar to an RTLola trigger. During execution we then iteratively update the database with new events. Crucially, the streaming database is optimized for such incremental computations and only updates the changed values in the query. This is a similar approach to the RTLola monitor, which also incrementally and efficiently updates its valuation upon encountering new events. We have generated seminar application scenarios with a varying number of applicants and report the average runtime of the three approaches in Figure~\ref{fig:DemoParity_Runtime}. We stopped at 500 applicants in the case of conditional statistical parity because the SQLite approach already took more than 100 seconds. The results show that RTLola is faster than the database approaches in our scenarios. As a side result, we also see that the streaming database RisingWave outperforms the SQLite implementation on all but the smallest inputs, which is even more pronounced for conditional statistical parity. Monitoring with RTLola still significantly outperforms even the advanced streaming database approach. This runtime advantage gets particularly important for systems meant to be deployed at a large scale, such as the COMPAS recidivism risk assessment tool.

\subsection{Monitoring Fairness of the COMPAS Tool}\label{sec:compas_eval}

\begin{figure}[t]
\centering
\begin{subfigure}{0.48\linewidth}
		\centering
\includegraphics[scale=0.37]{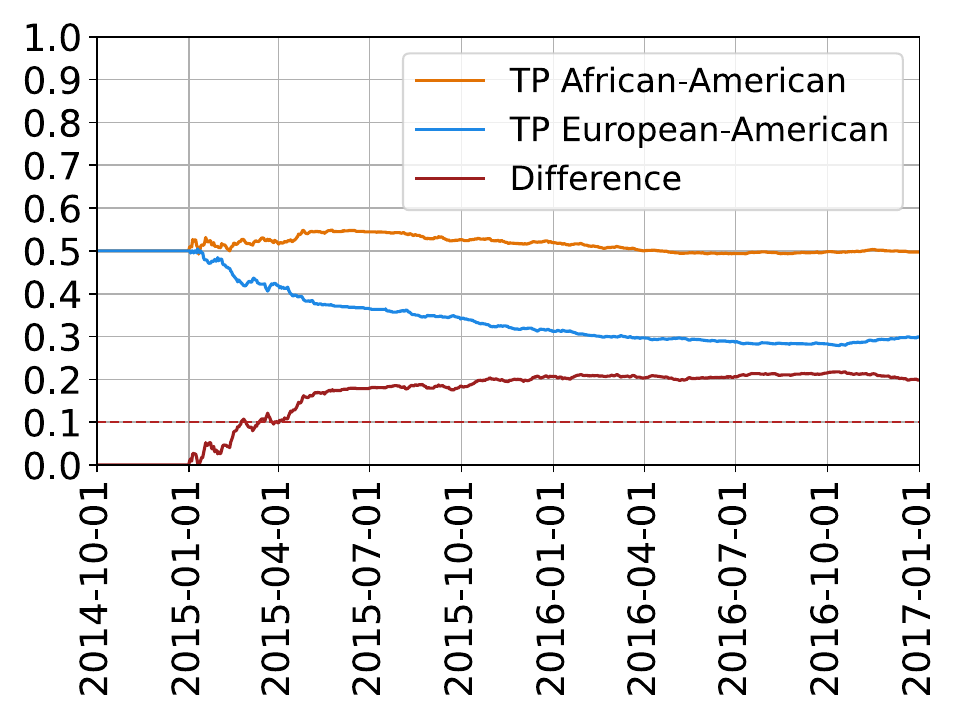}
\end{subfigure}\hspace{4pt}\begin{subfigure}{0.48\linewidth}
		\centering
\includegraphics[scale=0.37]{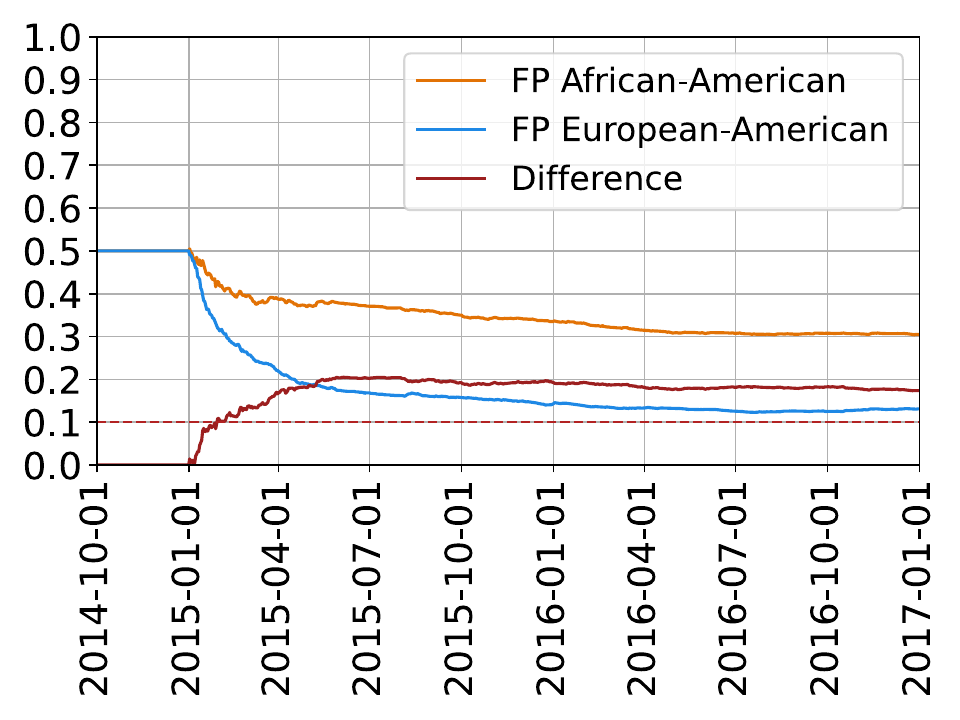}
\end{subfigure}

\caption{True positive rates (TP) and false positive rates (FP) of African-American and European-American defendants while monitoring equalized odds on the COMPAS data set~\cite{ProPublica16}. The dashed red line shows threshold $\epsilon$.}
\label{fig:COMPAS_Unfairness}
\end{figure}

We revisit the motivating example from Section~\ref{sec:motivation} to study the utility of our approach on real-world data from the recidivism risk prediction tool COMPAS. We use the same data set of COMPAS screenings between 2013 and 2014 in Broward County, Florida, which was also used by ProPublica~\cite{ProPublica16} in their original investigation. We converted their original data into streams that are temporally ordered to simulate online monitoring of the COMPAS tool. We then executed our RTLola monitor with the equalized-odds specification as outlined in Section~\ref{sec:implementation} for every combination of social groups. We used a confidence $\kappa$ of $100$, prior $\gamma(\omega) = 0.5$ and a threshold $\epsilon = 0.1$. In Figure~\ref{fig:COMPAS_Unfairness}, we illustrate the probability estimates and corresponding differences for African-American and European-American defendants. Note that the first two years are not shown, as a false positive result can only be definitely inferred after two years without recidivism, since this is the prediction horizon of the COMPAS tool as outlined in the COMPAS user guide~\cite{COMPAS}. As we can see, once the first two years have passed and the first definite outcomes can be inferred, unfairness can be established after less than a month, since the false positive rates of the groups quickly diverge. Since such tools are deployed over a long time-horizon, these initial two years without verdict bear comparatively little weight. Moreover, the judgment is robust and stays far above the threshold afterward. This experiment with data from the COMPAS tool shows that stream-based monitoring can be a viable method to detect unfairness of prediction systems early, and hence reduce the number of unfair decisions and predictions.

\section{Conclusion}

We have studied the monitoring of algorithmic fairness with the stream-based specification language RTLola. This language not only allows us to encode the estimation of conditional probabilities inherent to algorithmic-fairness specifications but also the timing requirements common to real-world applications where these specifications are crucial. We have demonstrated this exemplarily with the COMPAS tool that is used to predict the recidivism risk of defendants. Moreover, we have contributed a benchmark generator for constructing synthetic scenarios related to job application and university admission scenarios and have used these scenarios for an extensive evaluation of our approach, which shows that it is able to detect the ground truth reliably and efficiently. In the future, we plan on leveraging RTLola's innate capabilities for reasoning about data and time to express even more complex algorithmic-fairness specifications dealing with, e.g., expected values of credit scores or response times.

\subsubsection{Acknowledgments.}

This work was partially supported by the DFG in project 389792660 (TRR 248 -- CPEC) and by the ERC Grant HYPER (No. 101055412). Funded by the European Union. Views and opinions expressed are however those of the authors only and do not necessarily reflect those of the European Union or the European Research Council Executive Agency. Neither the European Union nor the granting authority can be held responsible for them.

\bibliographystyle{splncs04}
\bibliography{bibliography}
\end{document}

%% file: macros.tex
\newcommand{\prob}{\mathbb{P}}
\newcommand{\rtlola}{RTLola\xspace}

\newcommand{\ids}[0]{\texttt{ID}}
\newcommand{\idsI}[0]{\ids^\uparrow}
\newcommand{\idsO}[0]{\ids^\downarrow}

\newcommand{\srin}[0]{\idsI}
\newcommand{\srout}[0]{\idsO}
\newcommand{\sr}[0]{\srin \uplus \srout}

\newcommand{\StreamMap}[0]{\mathit{StreamMap}}

\newcommand{\instanceId}[0]{\mathit{InstanceID}}
\newcommand{\Stream}[0]{\mathit{Stream}}

\newcommand{\stime}[0]{\mathit{Time}}
\newcommand{\svalue}[0]{\mathbb{V}_{\bot}}

\newcommand{\world}[0]{\omega}
\newcommand{\World}[0]{\mathbb{W}}

\newcommand{\stimeMap}[0]{\mathit{TimeMap}}

\newcommand{\sid}[0]{\mathit{sid}}